\pdfoutput=1

\documentclass[11pt]{article}

\usepackage[final]{acl}

\usepackage{times}
\usepackage{latexsym}

\usepackage[T1]{fontenc}

\usepackage[utf8]{inputenc}

\usepackage{microtype}

\usepackage{inconsolata}

\usepackage{soul}
\usepackage{float}
\usepackage{capt-of}
\usepackage{enumitem}
\usepackage{amsmath,amssymb,amsbsy,amsfonts}
\usepackage{bbm}
\usepackage[ruled,linesnumbered]{algorithm2e}
\usepackage{booktabs,multirow,adjustbox,diagbox,threeparttable,tabularray,colortbl}
\usepackage{arydshln}
\usepackage{txfonts,bbm}
\usepackage{graphicx}
\usepackage[many]{tcolorbox}
\usepackage{xspace}
\usepackage{url}
\usepackage{cleveref}
\usepackage[symbol]{footmisc}

\definecolor{vgreen}{HTML}{60A917}

\title{\includegraphics[height=0.4cm,width=1.2cm]{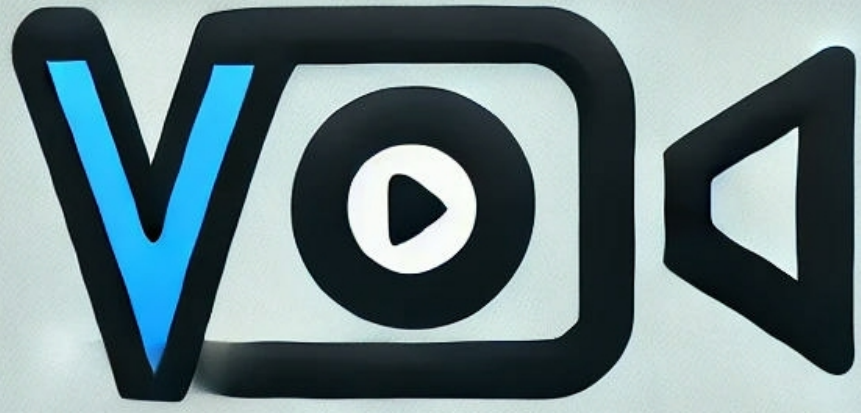} $\mathbbmss{ViBe}$: A Text-to-\ul{Vi}deo \ul{Be}nchmark for Evaluating Hallucination in Large Multimodal Models}

\author{
 \textbf{Vipula Rawte \textsuperscript{1}\thanks{Corresponding Author}},
 \textbf{Sarthak Jain\textsuperscript{2}\footnotemark[2]},
 \textbf{Aarush Sinha\textsuperscript{3}\footnotemark[2]},
 \textbf{Garv Kaushik\textsuperscript{4}\footnotemark[2]},
 \textbf{Aman Bansal\textsuperscript{5}\footnotemark[2]},
 \\
 \textbf{Prathiksha Rumale Vishwanath\textsuperscript{5}\footnotemark[2]},
 \textbf{Samyak Rajesh Jain\textsuperscript{6}},
 \textbf{Aishwarya Naresh Reganti\textsuperscript{7}\footnotemark[3]},
 \\
 \textbf{Vinija Jain\textsuperscript{8}\footnotemark[3]},
 \textbf{Aman Chadha\textsuperscript{9}\footnotemark[3]},
 \textbf{Amit Sheth\textsuperscript{1}},
 \textbf{Amitava Das\textsuperscript{1}}
\\
 \textsuperscript{1}AI Institute, University of South Carolina, USA,
 \textsuperscript{2}Guru Gobind Singh Indraprastha University, India,\\
 \textsuperscript{3}Vellore Institute of Technology, India,
 \textsuperscript{4}Indian Institute of Technology (BHU), India,\\
 \textsuperscript{5}University of Massachusetts Amherst, USA,
 \textsuperscript{6}University of California Santa Cruz, USA,\\
 \textsuperscript{7}Amazon Web Services,
 \textsuperscript{8}Meta,
 \textsuperscript{9}Amazon GenAI
}

\begin{document}
\maketitle

\footnotetext[2]{ Equal Contribution}
\footnotetext[3]{Worked independent of the position}

\begin{figure*}[!ht]
    \centering
    \includegraphics[width=1\textwidth]{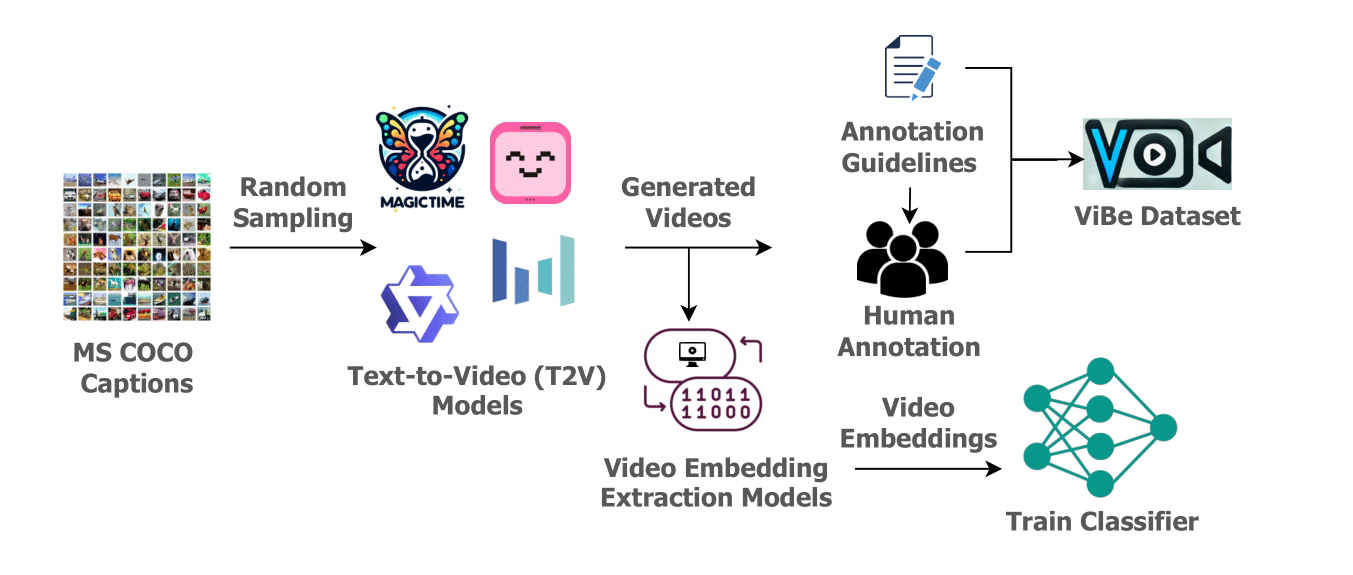}
    \caption{To generate the videos, we utilized randomly sampled image captions from the MS COCO dataset as textual inputs for the video generation models. The resulting videos were then manually annotated by human annotators to construct the ViBe dataset. Following annotation, the videos were processed into feature-rich video embeddings using advanced embedding techniques. These embeddings along with human annotated hallucination labels were subsequently input into various classifier models, which were trained to identify and categorize different types of video hallucinations, enabling the detection of discrepancies between the expected and generated content.}
    \label{pipe}
\end{figure*}

\begin{figure*}[!ht]
    \centering
    \includegraphics[width=1\linewidth]{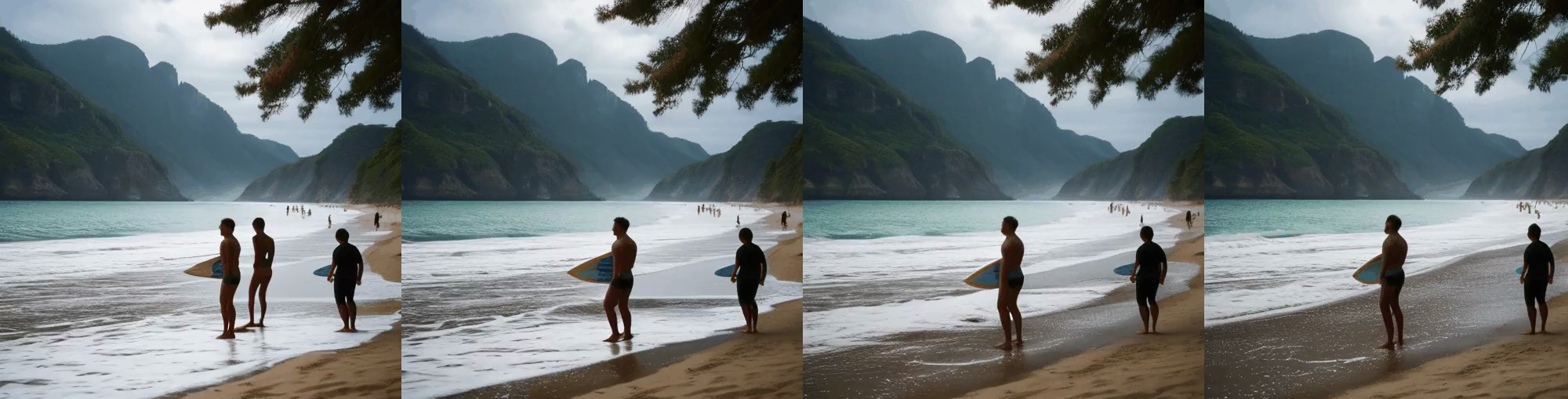}
    \caption{\textbf{Prompt:} three guys are standing on a beach next to surfboards. \textbf{\textcolor{red}{Vanishing Subject:}} The prompt mentions that there are three guys on a beach with surfboards. In the initial frame, we see 3 guys on the beach with surfboards, but in the last frame, we find only two guys remaining. The third guy seems to have \textit{vanished}.}
    \label{vs}
\end{figure*}

\begin{figure*}[!ht]
    \centering
    \includegraphics[width=0.8\linewidth]{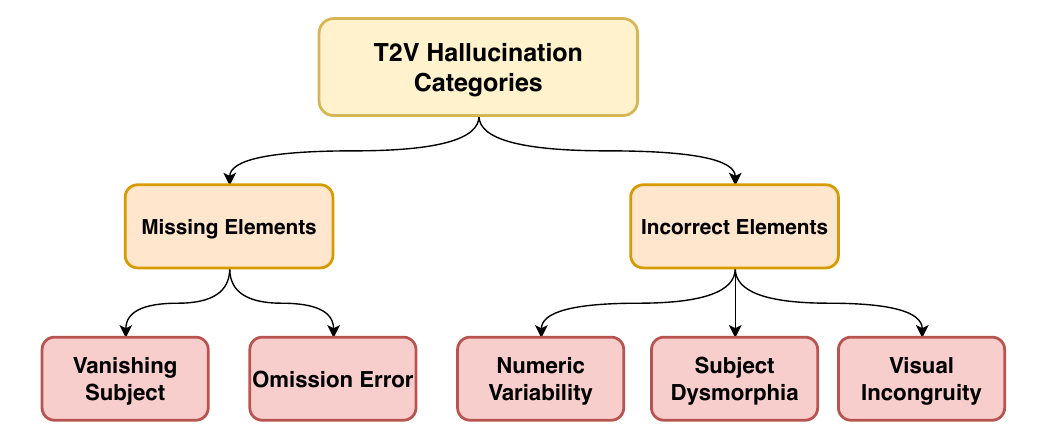} 
    \caption{Hierarchy of hallucination categories in $\mathbbmss{ViBe}$.}
    \label{fig:hallucination_hierarchy}
\end{figure*}

\begin{abstract}

Recent advances in Large Multimodal Models (LMMs) have expanded their capabilities to video understanding, with Text-to-Video (\texttt{T2V}) models excelling in generating videos from textual prompts. However, they still frequently produce hallucinated content, revealing AI-generated inconsistencies. We introduce $\mathbbmss{ViBe}$\footnote{\url{https://vibe-t2v-bench.github.io/}}: a large-scale dataset of hallucinated videos from open-source \texttt{T2V} models. We identify five major hallucination types: \textsc{Vanishing Subject}, \textsc{Omission Error}, \textsc{Numeric Variability}, \textsc{Subject Dysmorphia}, and \textsc{Visual Incongruity}. Using ten \texttt{T2V} models, we generated and manually annotated 3,782 videos from 837 diverse MS COCO captions. Our proposed benchmark includes a dataset of hallucinated videos and a classification framework using video embeddings. $\mathbbmss{ViBe}$ serves as a critical resource for evaluating \texttt{T2V} reliability and advancing hallucination detection. We establish classification as a baseline, with the TimeSFormer + CNN ensemble achieving the best performance (0.345 accuracy, 0.342 F1 score). While initial baselines proposed achieve modest accuracy, this highlights the difficulty of automated hallucination detection and the need for improved methods. Our research aims to drive the development of more robust \texttt{T2V} models and evaluate their outputs based on user preferences. 
\end{abstract}




\section{Introduction}
\label{sec:intro}
Text-to-video (\texttt{T2V}) models have advanced significantly, enabling the generation of coherent and visually detailed videos from textual prompts. These models have improved in capturing intricate visual elements that align with input text, yet a persistent challenge remains - the generation of hallucinated content. Hallucinations introduce visual discrepancies where elements either misalign with or distort the intended scene, compromising the realism and reliability of \texttt{T2V} outputs. This issue is particularly critical in applications that demand high fidelity to input prompts, such as content creation, education, and simulation systems.

To address this challenge, we introduce $\mathbbmss{ViBe}$, a comprehensive large-scale dataset designed to systematically analyze and categorize hallucinations in \texttt{T2V} models. Our dataset was constructed using 837 diverse captions from the MS COCO dataset, which were used to prompt 10 leading open-source \texttt{T2V} models, including HotShot-XL, MagicTime, AnimateDiff-MotionAdapter, and Zeroscope V2 XL. The resulting dataset consists of 3,782 videos, each manually annotated to identify common hallucination types, including disappearing subjects, missing scene components, numerical inconsistencies, and visual distortions.

$\mathbbmss{ViBe}$ serves as a valuable resource for evaluating the limitations of \texttt{T2V} models and facilitating the development of improved hallucination detection techniques. To complement the dataset, we propose a classification benchmark that leverages video embeddings from TimeSFormer and VideoMAE as inputs for hallucination classification. This benchmark establishes a structured evaluation pipeline, offering baseline performance results and highlighting the challenges of hallucination detection.

In summary, our key contributions are:

\begin{itemize}[leftmargin=25pt] 
    
    

    \item \textbf{A large-scale dataset for hallucination analysis in T2V models}: We introduce $\mathbbmss{ViBe}$, the first dataset focused on systematically categorizing hallucinations in generated videos. This dataset provides a foundation for studying errors in \texttt{T2V} generation and improving model fidelity.

    \item  \textbf{A structured framework for quantifying hallucinations}: We define five major hallucination categories and provide human-annotated labels, enabling researchers to analyze and mitigate common errors in \texttt{T2V} outputs.  

    \item  \textbf{A benchmark for hallucination classification}: We propose an evaluation framework using video embeddings and classification models to establish baseline performance on hallucination detection. Our results highlight the difficulty of this task and provide a reference for future improvements.  
        
\end{itemize}

\section{Related Work}

The phenomenon of hallucination in generative models has been widely studied across different types of media, including text, images, and videos. In text generation, large language models (LLMs) such as GPT-3 \cite{brown2020languagemodelsfewshotlearners} often produce responses that appear coherent but contain factual inaccuracies. This issue has motivated the development of evaluation benchmarks, such as the Hallucinations Leaderboard \cite{hallucinations-leaderboard}, which aim to measure how frequently and severely these models generate misleading or incorrect content.

In the case of image generation, models like DALL-E \cite{ramesh2022hierarchicaltextconditionalimagegeneration} and Imagen \cite{saharia2022photorealistictexttoimagediffusionmodels} have demonstrated impressive abilities in creating high-quality images from textual descriptions. However, these models sometimes generate artifacts that do not align with the provided input text, leading to unrealistic or misleading outputs. To address this problem, datasets such as the HAllucination DEtection dataSet (HADES) \cite{liu2022tokenlevelreferencefreehallucinationdetection} have been introduced. These datasets provide tools for assessing hallucination in text-to-image models by focusing on specific tokens and offering reference-free evaluation methods.

Video generation models face even greater challenges due to the added complexity of maintaining consistency across multiple frames. Errors in this context can manifest as unrealistic motion, sudden changes in object appearance, or scenes that contradict real-world physics. Recent efforts have aimed to detect and quantify hallucinations in text-to-video models (T2V). The Sora Detector \citep{chu2024soradetectorunifiedhallucination} is an example of a framework designed to identify hallucinations in video generation by analyzing key frames and comparing them against knowledge graphs. Similarly, VideoHallucer \citep{wang2024videohallucerevaluatingintrinsicextrinsic} introduces benchmarks to evaluate hallucinations by distinguishing between errors that originate from the model itself and those that arise due to external inconsistencies. Additionally, VBench \cite{huang2023vbench} provides a broad set of evaluation tools to assess the overall quality of generated videos.

Despite these advancements, a major limitation in current research is the lack of a large-scale, human-annotated dataset specifically designed to study hallucinations in text-to-video generation models. $\mathbbmss{ViBe}$ addresses this gap by introducing a structured large-scale dataset that categorizes different types of hallucinations observed in generated videos. This dataset includes a diverse collection of human-annotated videos sourced from ten publicly available T2V generative models. By providing detailed annotations, ViBe serves as a valuable resource for developing and testing new methods that aim to detect and reduce hallucinations in text-to-video models.

\section{Dataset Construction}

\subsection{Dataset Prompt Diversity}
To construct the $\mathbbmss{ViBe}$ dataset, we carefully selected 837 diverse captions from the MS COCO dataset \cite{lin2015microsoftcococommonobjects}, ensuring a balanced representation of real-world scenarios. These captions were used as prompts to generate 3,782 videos, making $\mathbbmss{ViBe}$ a valuable resource for evaluating text-to-video (\texttt{T2V}) models.

For structured evaluation, the dataset is organized into five distinct thematic categories:

\begin{itemize}
    \item \textbf{Sports}: This category includes prompts describing various athletic activities. An example caption is: \textit{"A baseball hitter stands in position to hit the ball."} These videos capture dynamic motion, human-object interactions, and fast-paced events.
    
    \item \textbf{Animals}: This category focuses on different species and their behaviors in natural and domestic settings. A sample prompt is: \textit{"Cows strain their necks for hay in between posts of a fence."} These videos challenge models to generate realistic animal motion and interactions with the environment.
    
    \item \textbf{Objects}: Prompts in this category describe static and dynamic objects in various contexts. For instance, \textit{"Two electrical boxes and signs sit on a street pole."} Evaluating this category helps analyze how well models capture object shapes, textures, and placements.
    
    \item \textbf{Environment and Settings}: This category includes prompts related to landscapes, weather conditions, and urban or rural scenes. An example caption is: \textit{"Two people in the distance on a beach with surfboards."} This set challenges models to generate coherent spatial layouts and realistic environmental details.
    
    \item \textbf{Human Activities}: This category involves prompts describing various actions performed by individuals or groups. For example, \textit{``Women are playing WII video games in a big room.''} The complexity of human movement, interactions, and physical realism is critical in evaluating these videos.
\end{itemize}

This structured approach ensures $\mathbbmss{ViBe}$ covers diverse real-world scenarios, spanning natural and urban environments, various human activities, and intricate object interactions. It enhances the dataset’s utility for evaluating the coherence and fidelity of generated videos while also serving as a foundation for benchmarking improvements in \texttt{T2V} model development.

\begin{figure*} [!ht]
    \centering
    \includegraphics[width=1\linewidth]{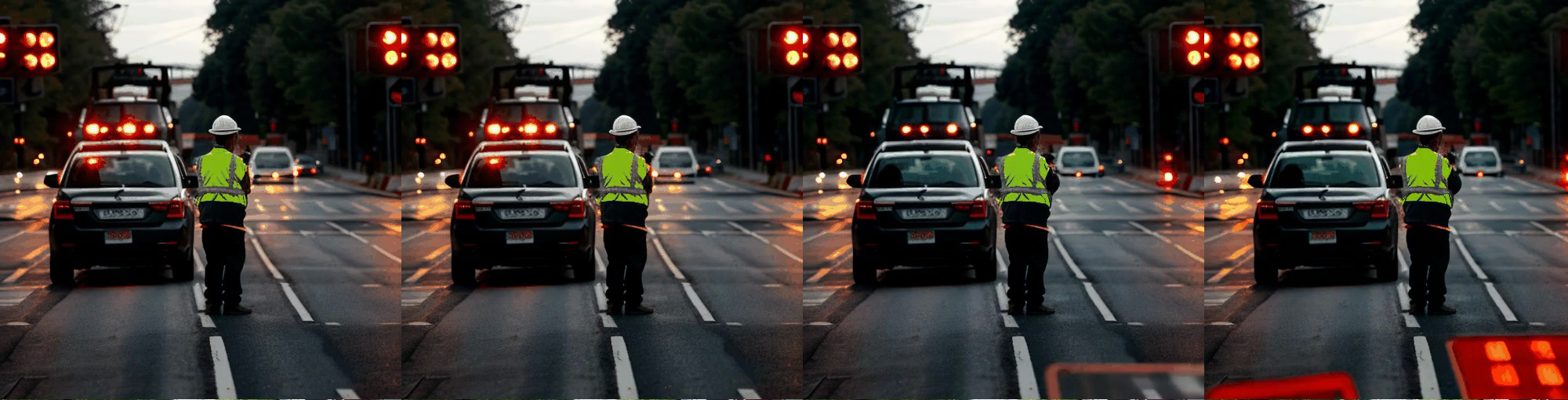}
    \caption{\textbf{Prompt:} Two road workers are standing by a red light with a sign. \textbf{\textcolor{red}{Numeric Variability:}} The prompt explicitly mentions two road workers. However, while the system accurately incorporates elements like the red light and depicts one road worker standing, it fails to generate the second road worker as specified in the prompt. The system modifies the specified number of subjects, decreasing their count, which deviates from the original instructions.}
    \label{smr}
\end{figure*}

\subsection{Models Used for Dataset Creation}
We used a subset of 837 captions as input prompts for ten \texttt{T2V} models, representing diverse architectures, sizes, and training paradigms: (i) MS1.7B \cite{ms-1.7b}, (ii) MagicTime \cite{yuan2024magictimetimelapsevideogeneration}, (iii) AnimateDiff-MotionAdapter \cite{animatediff-motion-adapter-v1-5-2}, (iv) zeroscope\_v2\_576w \cite{zeroscope_v2_576w}, (v) zeroscope\_v2\_XL \cite{zeroscope_v2_XL}, (vi) AnimateLCM \cite{animatelcm}, (vii) HotShotXL \cite{Mullan_Hotshot-XL_2023}, (viii) AnimateDiff Lightning \cite{lanimatedifflightning}, (ix) Show1 \cite{show1}, and (x) MORA \cite{mora}.

Most models generated 1-second videos, except Show1, which produced 2-second videos. Despite their brevity, the hallucination artifacts we define are highly discernible, enabling effective identification and analysis. Table \ref{tab:model_duration} provides a detailed breakdown of video duration across models, highlighting variability in generated outputs.

\begin{table}
\hfill
\centering
\footnotesize
\begin{tabular}{l|c}
\toprule
\textbf{{T2V} Model} & \textbf{{Duration}} \\
\midrule
\textbf{AnimateLCM} \cite{animatelcm} & 1 \\
\textbf{zeroscope\_v2\_XL} \cite{zeroscope_v2_XL} & 2 \\
\textbf{Show1} \cite{show1} & 2 \\
\textbf{MORA} \cite{mora} & 1 \\
\textbf{AnimateDiff Lightning} \cite{lanimatedifflightning} & 1 \\
\textbf{AnimateDiff-MotionAdapter} \cite{animatediff-motion-adapter-v1-5-2} & 1 \\
\textbf{MagicTime} \cite{yuan2024magictimetimelapsevideogeneration} & 1 \\
\textbf{zeroscope\_v2\_576w} \cite{zeroscope_v2_576w} & 2 \\
\textbf{MS1.7B} \cite{ms-1.7b} & 1 \\
\textbf{HotShotXL} \cite{Mullan_Hotshot-XL_2023} & 1 \\
\bottomrule
\end{tabular}
\caption{Video duration per model varies as follows: with the exception of the Show1 and ZeroscopeV2XL model, which generates videos with a duration of 2 seconds, all other models produce videos that are 1 second in length.}
\label{tab:model_duration}
\end{table}

Videos were systematically analyzed to identify and quantify hallucinations, revealing their widespread occurrence across various open-source \texttt{T2V} systems. Our dataset generation and classification benchmark pipeline are illustrated in Figure \ref{pipe}.

\begin{figure*}[!ht]
    \centering
    \includegraphics[width=1\linewidth]{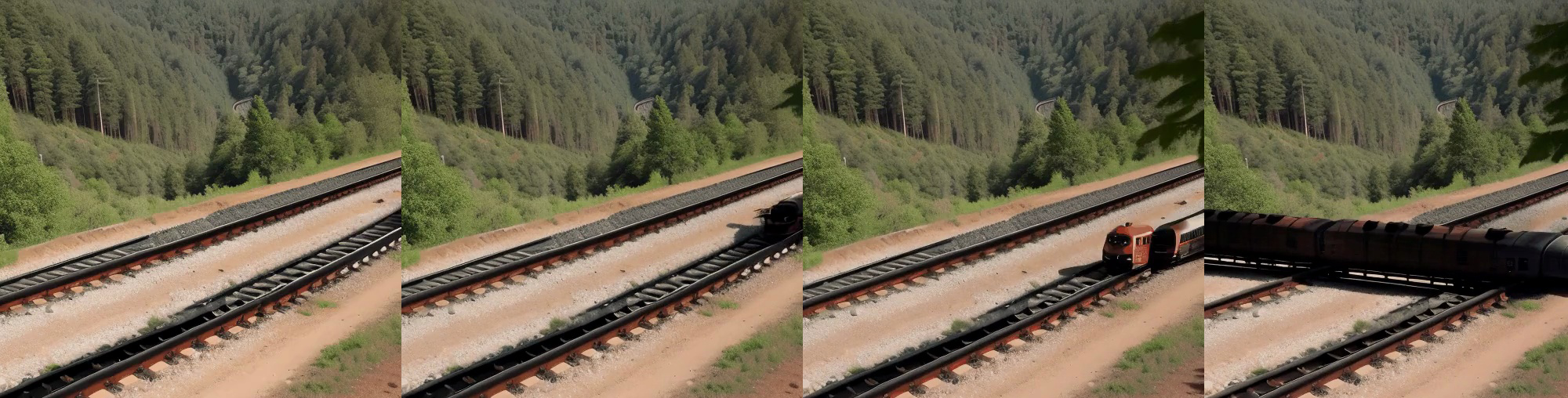}
    \caption{\textbf{Prompt:} A train heading for a curve in the track.
    \textbf{\textcolor{red}{Visual Incongruity:}} The scenario presents multiple logical and physical impossibilities in its temporal sequence. Initially, no train is visible in the first two frames, violating conservation of mass and the principle of object permanence. In the third frame, the train suddenly materializes on the track without a clear point of origin. In the final frame, the train inexplicably rotates to become perpendicular to the track, an action that defies both the mechanical constraints of train wheels on rails and basic laws of motion. This instantaneous 90-degree rotation would be physically impossible given a train's fixed wheel assembly and its momentum-governed movement along rails.}
    \label{pi}
\end{figure*}

\subsection{Hallucination Definitions}

\label{sec:hallucination_categories}
\begin{figure*}[!ht]
    \centering
    \includegraphics[width=1\linewidth]{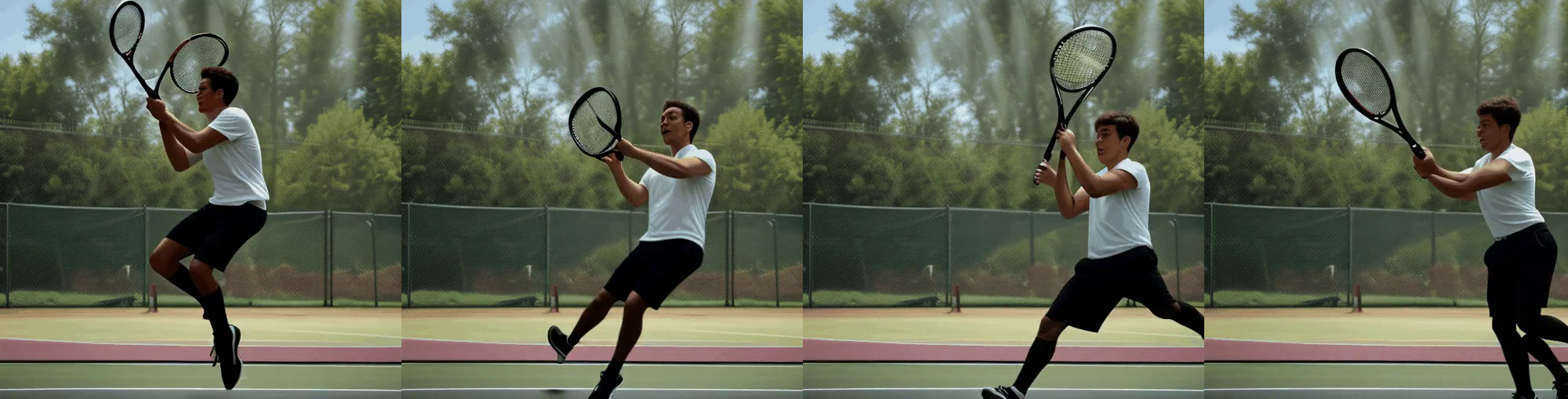}
    \caption{\textbf{Prompt:} A man in athletic wear swings a tennis racket through the air. \textbf{\textcolor{red}{{Subject Dysmorphia}:}} Throughout the video, both the man and the racket undergo visually inconsistent distortions, resulting in temporal and spatial anomalies. The system-generated artifacts introduce irregularities in the man’s form and the racket’s structure as they move, causing fluctuations in shape, scale, and position that disrupt the continuity of the intended action.}
    \label{tsd}
\end{figure*}

\begin{figure*}[!ht]
    \centering
    \includegraphics[width=1\linewidth]{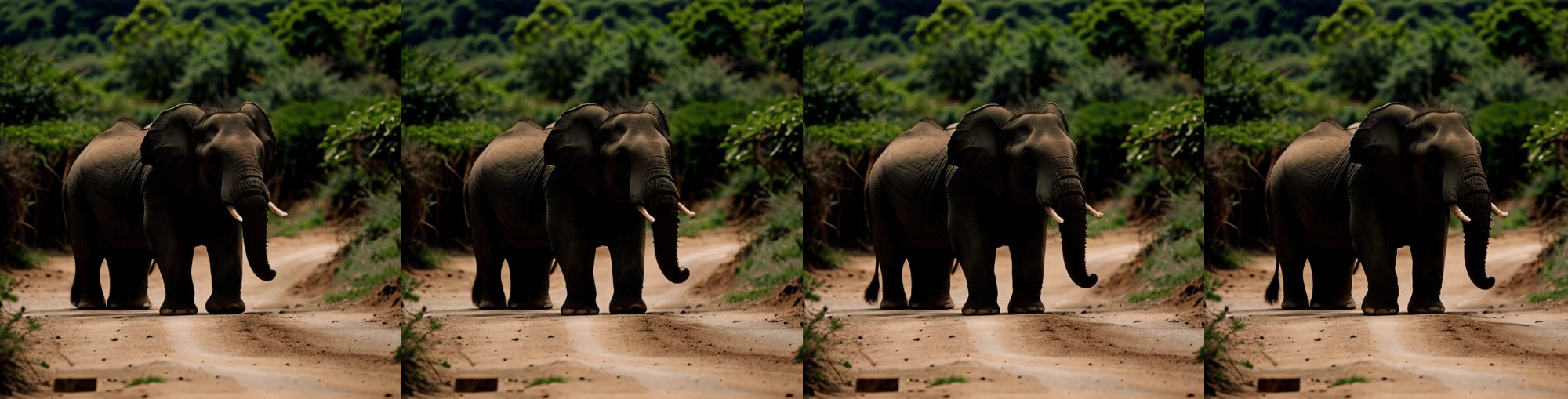}
    \caption{\textbf{Prompt:} a baby elephant walking behind a large one
    \textbf{\textcolor{red}{Omission Error:}} The generated output fails to render a critical component explicitly specified in the input prompt \textit{the larger one}. While the baby elephant is depicted, the absence of the larger elephant represents a significant deviation from the prompt requirements. This omission fundamentally alters the intended relationship and scale reference that was meant to be portrayed through the presence of both elephants, demonstrating incomplete prompt adherence.}
    \label{oe}
\end{figure*}

Hallucination categories were designed based on observed inconsistencies in generated videos rather than technical classifications like those in Sora Detector \citep{chu2024soradetectorunifiedhallucination}. These inconsistencies broadly fall into subject omissions or incorrect renderings, often exhibiting recurring patterns. We identified five distinct categories, which, while sometimes overlapping, are treated separately due to their frequent occurrence. This framework captures common hallucination patterns in \texttt{T2V} outputs, as detailed in the following section:

\begin{enumerate}

    \item \textbf{Vanishing Subject (VS):} A subject or part of a subject unpredictably disappears during the video. This is often observed in dynamic scenes where subjects fail to persist visually as seen in Figure \ref{vs}.

    \item \textbf{Omission Error (OE):} The video fails to render key elements explicitly described in the input prompt as seen in Figure \ref{oe}.
    
    \item \textbf{Numeric Variability (NV):} The video alters the specified number of subjects, either increasing or decreasing their count as seen in Figure \ref{smr}.
    
    \item \textbf{Subject Dysmorphia (SD):} Subjects in the video exhibit unnatural or distorted shapes, scales, or orientation changes, violating expected physical consistency during the course of the video as seen in Figure \ref{tsd}. 
    
    \item \textbf{Visual Incongruity (VI):} Logically incompatible or physically impossible elements are combined, creating perceptual inconsistencies or violating natural laws as seen in Figure \ref{pi}.

\end{enumerate}

\subsection{Human Annotation Details}
Table \ref{tab:model_comparison} presents the distribution of hallucinated videos across models and categories. Five annotators manually categorized 3,782 videos, assigning each to the most prominent hallucination type based on a predefined taxonomy. To ensure consistency, they followed a hierarchical classification approach, prioritizing specific sub-categories before broader ones. Figure \ref{fig:hallucination_hierarchy} visually represents this hierarchy. Additional details on dataset annotation are provided in the \cref{sec:appendix}.

\subsection{Implementation Details} \label{sec:imp}

For embedding extraction and classifier training, the process utilized a system with 8 CPU cores, each equipped with 32 GB of memory. This hardware configuration provided the necessary computational resources to efficiently handle data processing and model training. For video generation tasks, an NVIDIA A100 GPU \cite{9361255} was employed, taking advantage of its high-performance capabilities for accelerated computation and rendering of complex video content.

\begin{figure*}[!ht]
    \centering
    \begin{minipage}{0.48\textwidth}
        \centering
        \includegraphics[width=\linewidth]{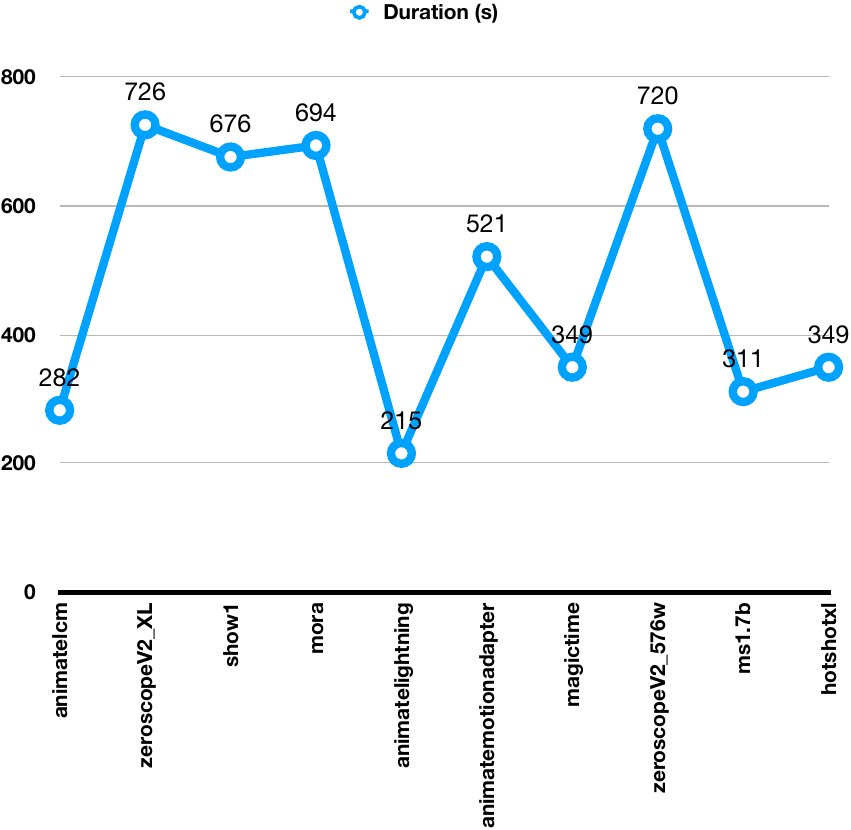}
        \caption{The total duration per model represents the cumulative duration of all videos associated with that model. For instance, \textbf{magictime} has a cumulative video duration of 349 seconds. The total duration for \textbf{zeroscopeV2\_XL} has the longest time, with a duration of 726 seconds, followed by \textbf{zeroscopeV2\_576w} at 720 seconds. In contrast, the shortest time was recorded for \textbf{animatelightning}, which took 215 seconds.}
        \label{fig:duration}
    \end{minipage}\hfill
    \begin{minipage}{0.48\textwidth}
        \centering
        \includegraphics[width=\linewidth]{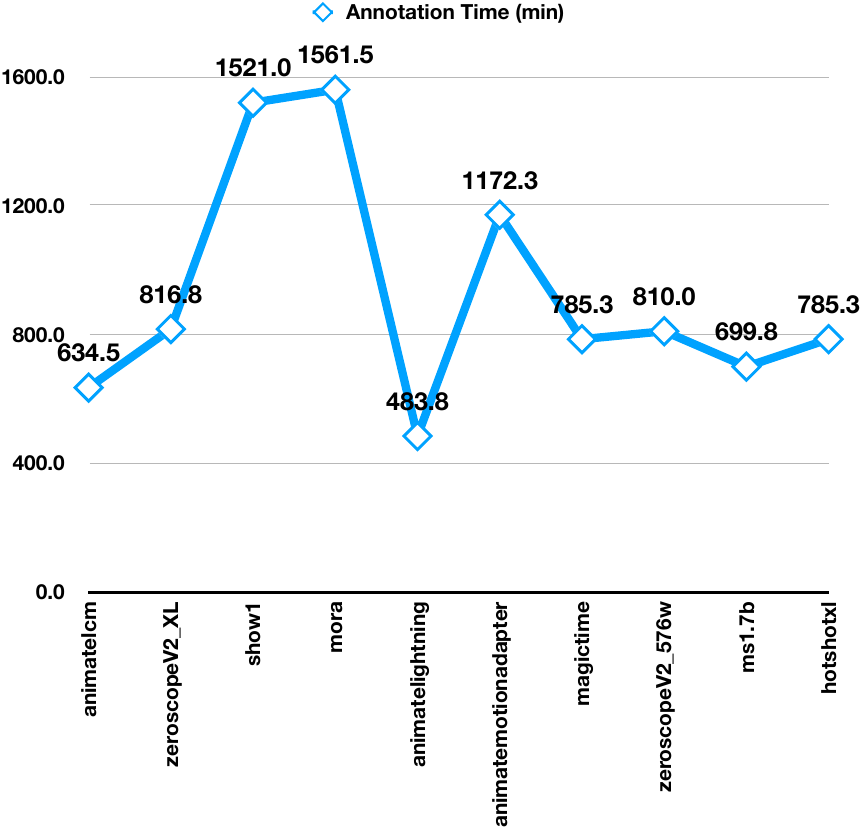}
        \caption{The longest annotation time was recorded for \textbf{mora}, taking 1561.5 minutes, followed by \textbf{show1} at 1521.0 minutes. Conversely, the shortest annotation time was observed for \textbf{animatelightning}, which required 483.75 minutes.}
        \label{fig:annot_time}
    \end{minipage}
\end{figure*}

The total duration per model refers to the cumulative time spent annotating all videos associated with that specific model, as shown in \ref{fig:duration}. \ref{tab:model_duration} provides a detailed report on the video length for each model, allowing for an analysis of how video duration may impact processing times or model performance during annotation tasks.

\subsection{Inter-Annotator Scores}
Two annotators were given 100 common videos to assess inter-annotator agreement, compared against the dataset's gold-standard annotations. Cohen’s Kappa scores (Table \ref{tab:agreement_metrics}) show the highest agreement for Visual Incongruity (0.8737) and the lowest for Omission Error (0.7474). Cohen’s Kappa is calculated as:
\begin{equation}
\kappa = \frac{p_o - p_e}{1 - p_e}
\end{equation}

where:
\begin{itemize}
    \item \( p_o \) is the observed agreement between the raters.
    \item \( p_e \) is the expected agreement by chance.
\end{itemize}


\begin{table*}[!htb]
\centering
\begin{minipage}{\columnwidth}
\centering
\resizebox{\columnwidth}{!}{%
\begin{tabular}{lcccccc}
\toprule
\texttt{T2V} Model & \textbf{VS} & \textbf{NV} & \textbf{SD} & \textbf{OE} & \textbf{VI} & \textbf{Total} \\
\toprule
\textbf{AnimateLCM} & 2 & 70 & 70 & 70 & 70 & 282 \\
\textbf{zeroscope\_v2\_XL} & 18 & 0 & 37 & 109 & 199 & 363 \\
\textbf{Show1} & 13 & 71 & 88 & 111 & 55 & 338 \\
\textbf{MORA} & 82 & 96 & 99 & 202 & 215 & 694 \\
\textbf{AnimateDiff Lightning} & 11 & 33 & 52 & 56 & 63 & 215 \\
\textbf{AnimateDiff-MotionAdapter} & 28 & 59 & 158 & 182 & 94 & 521 \\
\textbf{MagicTime} & 70 & 70 & 70 & 69 & 70 & 349 \\
\textbf{zeroscope\_v2\_576w} & 17 & 0 & 41 & 115 & 187 & 360 \\
\textbf{MS1.7B} & 51 & 50 & 70 & 70 & 70 & 311 \\
\textbf{HotShotXL} & 70 & 70 & 70 & 69 & 70 & 349 \\
\midrule
\textbf{Total} & 362 & 519 & 755 & 1053 & 1093 & 3782 \\
\bottomrule
\end{tabular}
}
\caption{This table shows the distribution of hallucinated videos produced by ten different text-to-video models, classified into five types of hallucinations. The dataset includes 3,782 videos, each assessed for the occurrence of these hallucination types.}
\label{tab:model_comparison}
\end{minipage}
\hfill
\begin{minipage}{\columnwidth}
\centering
\scriptsize
\resizebox{\textwidth}{!}{%
\begin{tabular}{lcc}
\toprule
\textbf{Hallucination Categories} & \textbf{Cohen's Kappa}\\
\toprule
\textbf{Vanishing Subject} & 0.7660 \\
\textbf{Omission Error} & 0.7474 \\
\textbf{Numeric Variability} & 0.8500 \\
\textbf{Subject Dysmorphia} & 0.8173 \\
\textbf{Visual Incongruity} & 0.8737 \\
\bottomrule
\end{tabular}
}
\caption{This table presents Cohen's Kappa Score for Evaluating Inter-Annotator Agreement. The score ranges from \textbf{-1 to 1}: \textbf{1} represents perfect agreement between annotators. \textbf{0} implies that the agreement is no better than random chance. \textbf{Negative values} indicate stronger disagreement than expected by chance, suggesting systematic annotation inconsistencies.}
\label{tab:agreement_metrics}
\end{minipage}
\end{table*}

\section{Classification}
\label{sec:benchmark}

Given the growing challenge of video hallucinations, addressing this issue is crucial. Currently, the literature includes only one \texttt{T2V} hallucination benchmark, T2VHaluBench \cite{chu2024soradetectorunifiedhallucination}, which consists of just 50 videos, limiting its utility for robust evaluation. To overcome this, we propose a large dataset to drive further research, along with several classical classification baselines to support hallucination category prediction. We expect this work to be a key resource for advancing research in this domain.

\subsection{\texttt{T2V} Hallucination Classification}

We evaluate our $\mathbbmss{ViBe}$ dataset using a variety of classification models. We also present a novel task for classifying hallucinations in a text-to-video generation. The first step involves extracting video embeddings from two pre-trained models: VideoMAE (Video Masked Autoencoders for Data-Efficient Pretraining) \cite{tong2022videomaemaskedautoencodersdataefficient} and TimeSFormer (Time-Space Attention Network for Video Understanding) \cite{bertasius2021spacetimeattentionneedvideo}. These extracted embeddings are subsequently used as feature representations for seven distinct classification algorithms: Long Short-Term Memory (LSTM) \cite{NIPS2014_a14ac55a}, Transformer \cite{NIPS2017_3f5ee243}, Convolutional Neural Network (CNN) \cite{NIPS2012_c399862d}, Gated Recurrent Unit (GRU) \cite{chung2014empirical}, Recurrent Neural Network (RNN) \cite{mikolov10_interspeech}, Random Forest (RF) \cite{ho1995random}, and Support Vector Machine (SVM) \cite{cortes1995support}. This comprehensive evaluation across different model architectures allows for a thorough comparison of performance in classifying the given video dataset.

\begin{table}[H]
\footnotesize
\centering
\begin{tabular}{lc}
\toprule
\textbf{\texttt{T2V} Hallucination Benchmark} & \textbf{\# Videos}  \\
\toprule
\textbf{T2VHaluBench} \cite{chu2024sora} & 50  \\
$\mathbbmss{ViBe}$ & 3,782  \\
\bottomrule
\end{tabular}\caption{The current \texttt{T2V} Hallucination Benchmark, T2VHaluBench, is limited by a small sample size in its dataset. In contrast, our dataset significantly outpaces it, comprising a substantial collection of 3,782 videos, offering a more comprehensive and robust foundation for evaluating \texttt{T2V} hallucination phenomena.}
\label{tab:T2V_bench}
\end{table}

\subsection{Experimental Setup}

The dataset was partitioned into 80\% for training and 20\% for testing, and the Adam/AdamW optimizer was used \cite{loshchilov2018decoupled}..

\begin{table}[!htp]\centering
\scriptsize
\resizebox{\columnwidth}{!}{%
\begin{tabular}{lccccr}\toprule
\textbf{} &\multicolumn{4}{c}{\textbf{Hyperparameters}} \\\cmidrule{2-5}
\textbf{Model} &\textbf{\# Epochs} &\textbf{Batch size} &\textbf{Optimizer} &\textbf{Loss} \\\midrule
\textbf{GRU} &30 &32 &AdamW &categorical\_crossentropy \\
\textbf{LSTM} &120 &128 &Adam &categorical\_crossentropy \\
\textbf{Transformer} &100 &128 &Adam &categorical\_crossentropy \\
\textbf{CNN} &100 &128 &Adam &categorical\_crossentropy \\
\textbf{RNN} &120 &128 &Adam &categorical\_crossentropy \\
\textbf{RF} &\multicolumn{4}{c}{N/A} \\
\textbf{SVM} &\multicolumn{4}{c}{N/A} \\
\bottomrule
\end{tabular}%
}
\caption{Specifications of the model hyperparameters employed during the classifier training process: for both RF and SVM classifiers, default settings from scikit-learn \cite{scikit-learn} were applied.}
\label{tab:hyperp}
\end{table}


\begin{table}[!ht]
    \centering
    \footnotesize
    \renewcommand{\arraystretch}{1.2}
    \resizebox{\columnwidth}{!}{%
    \begin{tabular}{lcc}
        \toprule
        \textbf{Model} & \textbf{Accuracy $\uparrow$} & \textbf{F1 Score $\uparrow$} \\
        \toprule
        \textbf{VideoMAE + GRU} & 0.268 & 0.190 \\
        \textbf{VideoMAE + LSTM} & 0.302 & 0.299 \\
        \textbf{VideoMAE + Transformer} & 0.284 & 0.254 \\
        \textbf{VideoMAE + CNN} & 0.303 & 0.290 \\
        \textbf{VideoMAE + RNN} & 0.289 & 0.289 \\
        \textbf{VideoMAE + RF}  & 0.331 & 0.279 \\
        \textbf{VideoMAE + SVM}  & 0.277 & 0.282 \\
        \hdashline
        \textbf{TimeSFormer + GRU} & 0.325 & 0.279 \\
        \textbf{TimeSFormer + LSTM} & 0.337 & 0.334 \\
        \textbf{TimeSFormer + Transformer} & 0.322 & 0.284 \\
        \textbf{TimeSFormer + CNN} & \colorbox{green!30}{\textbf{0.345}} & \colorbox{green!30}{\textbf{0.342}} \\
        \textbf{TimeSFormer + RNN} & 0.299 & 0.299 \\
        \textbf{TimeSFormer + RF}  & 0.341 & 0.282 \\
        \textbf{TimeSFormer + SVM}  & 0.270 & 0.274 \\
        \bottomrule
    \end{tabular}
    }
    \caption{A detailed comparison of model accuracy and F1 score is presented for various combinations of models utilizing VideoMAE and TimeSFormer embeddings. The model yielding the highest performance is denoted in \colorbox{green!30}{\textbf{green}} for easy identification. This analysis aims to assess the effectiveness of different embedding strategies in optimizing both classification accuracy and the balance between precision and recall, as captured by the F1 score.}
    \label{tab:model_performance}
    \vspace{-5mm}

\end{table}

For classification, video embeddings were extracted using the TimeSformer and VideoMAE models, which process individual frames to generate meaningful feature representations. However, despite these models operating on a per-frame basis, the classification task itself did not strictly follow a frame-by-frame approach. Instead, the classification was performed at a higher level, incorporating aggregated representations of the extracted embeddings. 

\subsection{Results and Analysis}

Table \ref{tab:model_performance} presents a comprehensive comparison of the performance metrics, namely accuracy and F1 score, for each model across two distinct feature sets: VideoMAE and TimeSFormer embeddings. 

For the models trained with VideoMAE embeddings, the RF model demonstrated the highest accuracy, achieving a value of 0.331. However, the LSTM model excelled in the F1 score, recording the highest value of 0.299. On the other hand, the GRU model exhibited the lowest performance, with an accuracy of 0.268 and an F1 score of 0.190, indicating a significant drop in both metrics compared to the other models in this category.

When the TimeSFormer embeddings were utilized, the CNN model outperformed all other models, attaining both the highest accuracy (0.345) and F1 score (0.342). The LSTM model also performed competitively, yielding an accuracy of 0.337 and an F1 score of 0.334. In contrast, the SVM model was the least effective, with an accuracy of 0.270 and an F1 score of 0.274, which were notably lower than those of other models.

Overall, TimeSFormer embeddings consistently outperformed VideoMAE embeddings across most models, showing superior accuracy and F1 scores. The combination of TimeSFormer embeddings with the CNN model delivered the optimal performance in terms of both accuracy and F1 score, making it the most effective configuration in this study.

\section{Conclusion and Future Work}




In this paper, we present ViBe, a large-scale dataset of 3,782 manually annotated videos, surpassing prior benchmarks like T2VHaluBench by 75 times in scale. It provides a robust foundation for evaluating hallucination, ensuring prompt adherence, and improving video generation quality across diverse scenarios across \texttt{T2V} models. We introduce a five-category hallucination taxonomy, enabling systematic analysis and benchmarking of \texttt{T2V} models. 

Future research directions encompass several key areas of improvement. First, expanding the existing taxonomy will provide a more comprehensive framework for categorizing and understanding various aspects of video generation. Additionally, evaluating longer-duration videos will help assess the scalability and temporal coherence of the models over extended sequences. Another critical focus is the development of automated classification techniques, which will enhance the efficiency and accuracy of video analysis by reducing reliance on manual annotation. Finally, an essential step forward involves training \texttt{T2V} models using RLHF. This approach aims to refine the alignment of generated videos with human preferences, improving the synthesized content's relevance and quality.

\section{Limitations}
ViBe, while robust, has some limitations. Videos are classified into a single hallucination category for streamlined annotation, which may overlook multi-category overlaps. The dataset is also limited to short video durations due to constraints in open-source \texttt{T2V} models and annotation feasibility. Future work could address these limitations by incorporating multi-category annotations and extending video durations as computational and automatic annotation methods improve.

\section{Ethics Statement}
Our research on the video hallucinations benchmark aims to advance the understanding and evaluation of generative models, ensuring transparency and accountability in their development. We acknowledge the ethical concerns surrounding potential misuse, particularly in creating highly realistic doctored videos that could contribute to misinformation, fraud, or manipulation. To mitigate these risks, we emphasize responsible disclosure, promote the use of our benchmark for detection and mitigation efforts, and advocate for ethical AI development practices. 

\bibliography{acl_latex}

\newpage
\appendix
\section{Appendix} \label{sec:appendix}

This section offers supplementary material, including additional examples, implementation details, and more, to enhance the reader's understanding of the concepts discussed in this work. We also present additional details of the annotation process in Section \ref{sec:annot}.

\section{Annotation Details} \label{sec:annot}

The objective of this annotation task is to detect and classify hallucinations in videos produced by \texttt{T2V} models. The annotated data will be utilized to assess the model’s adherence to input prompts and contribute to improving hallucination detection and mitigation.

\begin{figure*}[!ht]
    \centering
    \includegraphics[width=\linewidth]{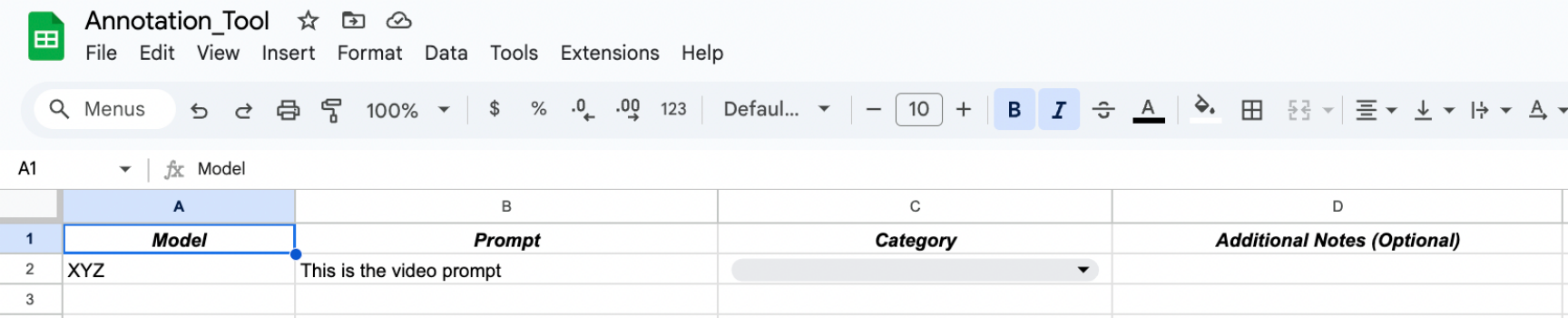}
    \caption{This figure illustrates the annotation tool employed to label various video samples. The tool comprises four columns: \\
    \textbf{Model:} Represents the specific T2V model. \\
    \textbf{Prompt:} Contains the image caption text derived from the MS COCO dataset. \\
    \textbf{Category:} Indicates one of the five predefined hallucination categories. \\
    \textbf{Additional Notes:} An optional column for supplementary annotations.}
    \label{fig:annot1}
\end{figure*}

\begin{enumerate}[label=\textbf{\arabic*}, left=0.5em]
    \item \textbf{Understanding Hallucination Categories} Annotators will be trained to recognize the five predefined categories of \texttt{T2V} hallucination: \texttt{Vanishing Subject}, \texttt{Omission Error}, \texttt{Numeric Variability}, \texttt{Subject Dysmorphia}, = and \texttt{Visual Incongruity}. 
    
    \item \textbf{Training and Evaluation Protocol}  
        
        \begin{enumerate}[label=\textbf{\alph*.}]
        \item \textbf{Training:} Annotators will receive example videos for each hallucination category, along with justifications for category assignments.  

        \item \textbf{Evaluation:} Annotators will classify five test videos, each corresponding to a unique hallucination category. A minimum agreement score of 60\% (correct classification of at least 3 out of 5 videos) is required to proceed to the annotation phase.  

        \item \textbf{Feedback Loop:} Annotators who do not meet the agreement threshold will receive targeted feedback and additional training.  
        \end{enumerate}

    \item \textbf{Annotation Process}
        \begin{enumerate}[label=\textbf{\alph*.}]
        \item \textbf{Video Evaluation:} Annotators will carefully review the assigned video, comparing the visual content to the input text prompt to identify inconsistencies.  

        \item \textbf{Hallucination Categorization:} Annotators will assign the most prominent hallucination category. If multiple hallucinations exist, the most visibly apparent one will be selected.  

        \item \textbf{Annotation Tool:} The identified category will be entered into the annotation tool (see \ref{fig:annot1}, \ref{fig:annot2}). Supplementary notes can be added for clarification.  

        \item \textbf{Annotation Time:} The average annotation time was recorded at 2.25 seconds per video (see \ref{fig:annot_time}).
        \end{enumerate}
\end{enumerate}

\begin{figure*}[!ht]
    \centering
    \includegraphics[width=\linewidth]{images/annot2.pdf}
    \caption{Using this annotation tool, annotators can classify the generated videos into one of the five predefined hallucination categories.}
    \label{fig:annot2}
\end{figure*}

\section{Dataset} \label{sec:dataset}

The five categories of hallucination have been previously defined, with examples provided for each. In this section, we will present additional examples to further illustrate these categories.

\subsection{Hallucination Categories}

\begin{figure*}
    \centering
    \includegraphics[width=\linewidth]{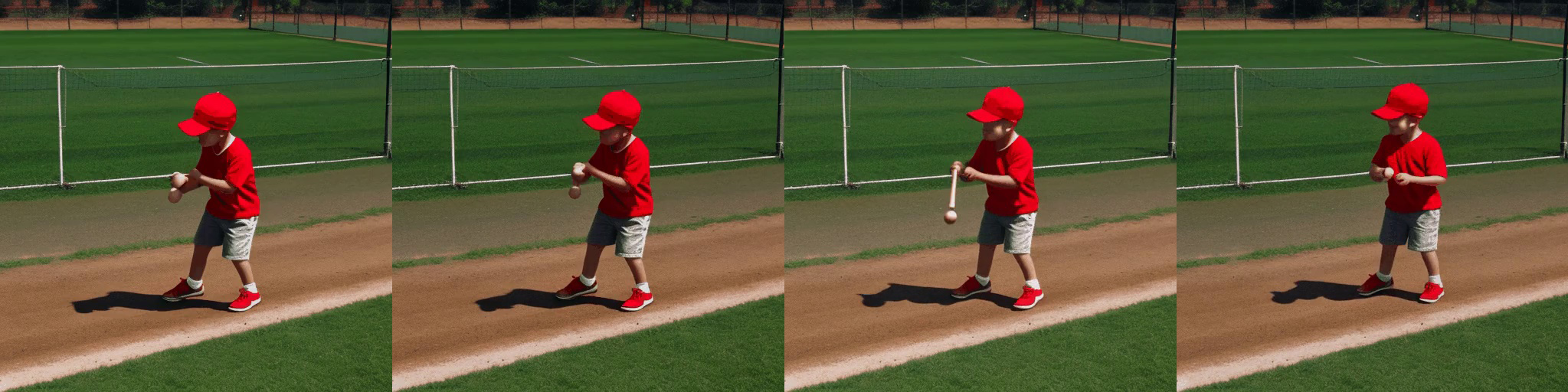}
    \caption{\textbf{Prompt:} A boy in a red hat playing with tee ball set. \textbf{\textcolor{red}{{Vanishing Subject}:}} The visual content depicts a boy wearing a red hat engaged in play with a tee-ball set. However, a hallucination occurs within the generated scene, where the tee-ball set, initially present, inexplicably disappears during the sequence.}
    \label{fig:vs1}
\end{figure*}

\begin{figure*}
    \centering
    \includegraphics[width=\linewidth]{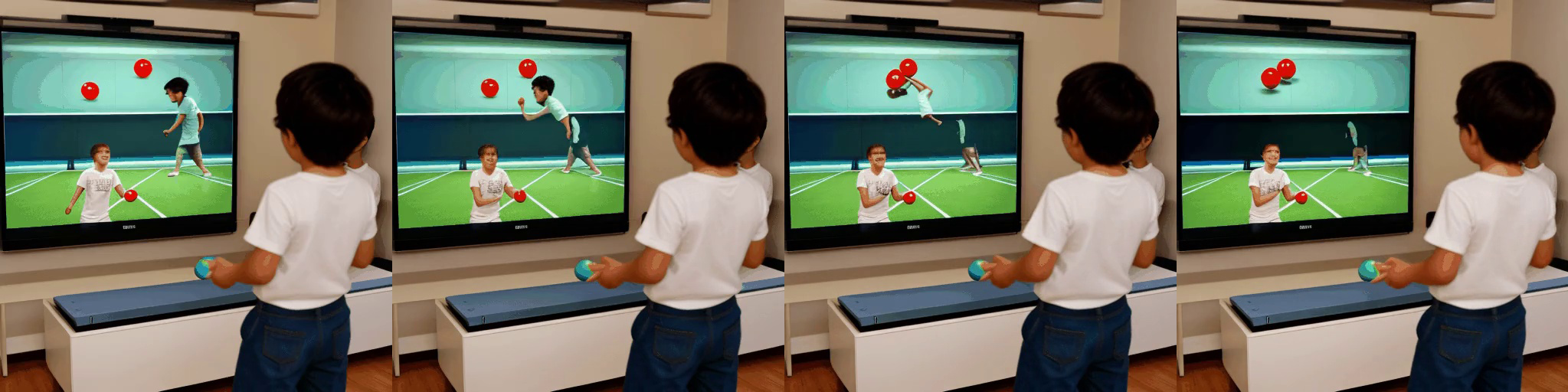}
    \caption{\textbf{Prompt:} Two young boys playing Wii bowling on a large television screen \textbf{\textcolor{red}{{Vanishing Subject}:}} In the video frames, the TV initially displays two boys. However, as the video progresses, subtle changes occur. By the final frame, one of the boys on the TV has mysteriously vanished, leaving only the other behind.}
    \label{fig:vs2}
\end{figure*}

\begin{figure*}
    \centering
    \includegraphics[width=\linewidth]{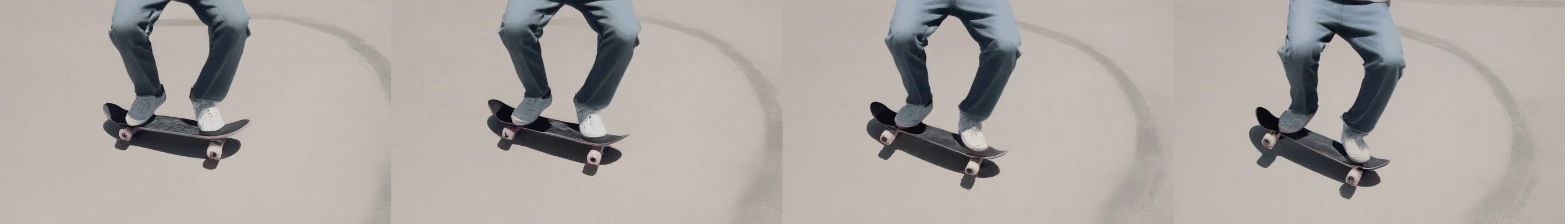}
    \caption{\textbf{Prompt:} A person on a skateboard with his arms in the air. \textbf{\textcolor{red}{{Omission Error}:}} The prompt describes a scene featuring a person on a skateboard with their arms raised in the air. However, this description exhibits a hallucination, as the video does not depict the individual's arms at all.}
    \label{fig:oe1}
\end{figure*}

\begin{figure*}
    \centering
    \includegraphics[width=\linewidth]{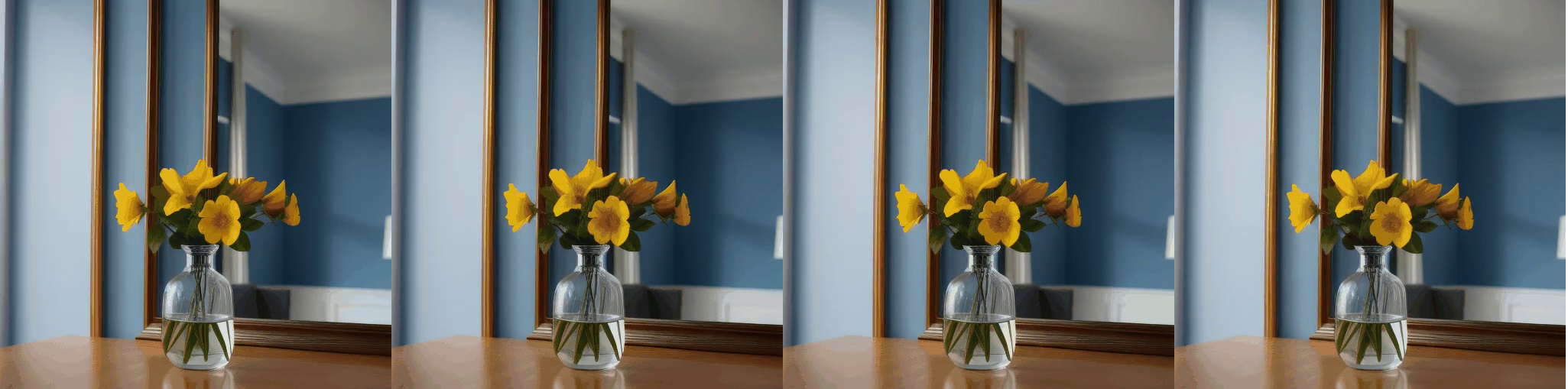}
    \caption{\textbf{Prompt:} Blue and yellow flowers in a glass vase near a mirror. \textbf{\textcolor{red}{{Omission Error}:}} The video lacks any blue flowers, despite their explicit mention in the prompt. This discrepancy highlights a failure of the model to accurately represent key visual elements specified in the input.}
    \label{fig:oe2}
\end{figure*}

\begin{figure*}
    \centering
    \includegraphics[width=\linewidth]{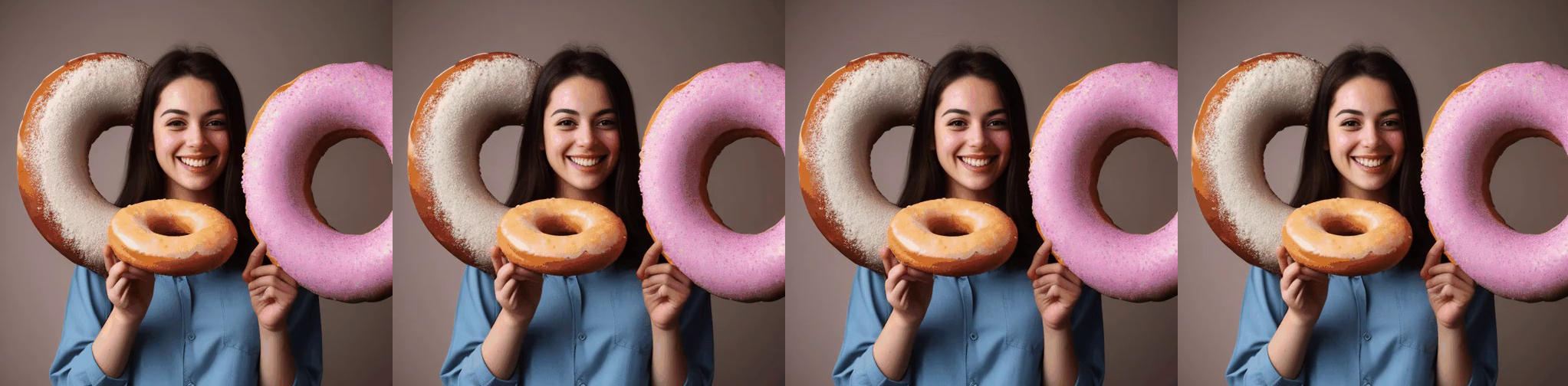}
    \caption{\textbf{Prompt:} A happy adult holding two large donuts. \textbf{\textcolor{red}{{Numeric Variability}:}} The description depicts a content scenario where a happy adult is holding two large donuts. However, a hallucination occurs within the video, where the depicted woman is shown holding three donuts instead of two.}
    \label{fig:nv1}
\end{figure*}

\begin{figure*}
    \centering
    \includegraphics[width=\linewidth]{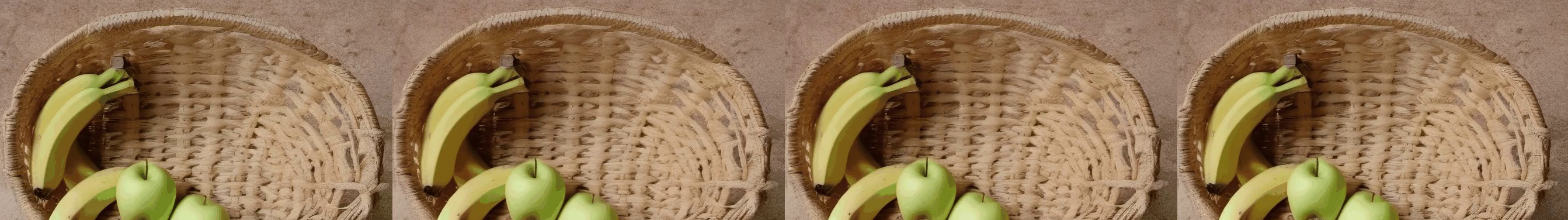}
    \caption{\textbf{Prompt:} A banana and a yellow apple in a woven basket. \textbf{\textcolor{red}{{Numeric Variability}:}} The visual scene consists of a woven basket containing one banana and one yellow apple. However, the generative output exhibits a hallucination, inaccurately depicting two bananas and two apples within the basket.}
    \label{fig:nv2}
\end{figure*}

\begin{figure*}
    \centering
    \includegraphics[width=\linewidth]{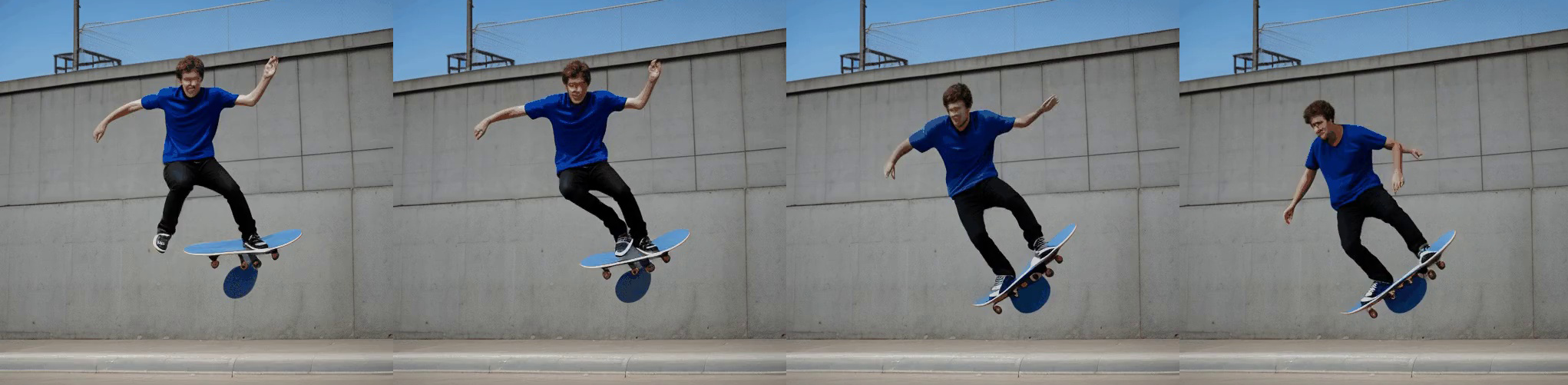}
    \caption{\textbf{Prompt:} Skateboarder and blue shirt and black jeans jumping on his board \textbf{\textcolor{red}{{Subject Dysmorphia}:}} The video depicts a person riding a skateboard. Throughout the frames, the wheels of the skateboard keep morphing, fluctuating in number as they increase and decrease. Additionally, the skateboarder’s arms undergo a similar distortion, gradually shifting in shape over time.}
    \label{fig:td1}
\end{figure*}

\begin{figure*}
    \centering
    \includegraphics[width=\linewidth]{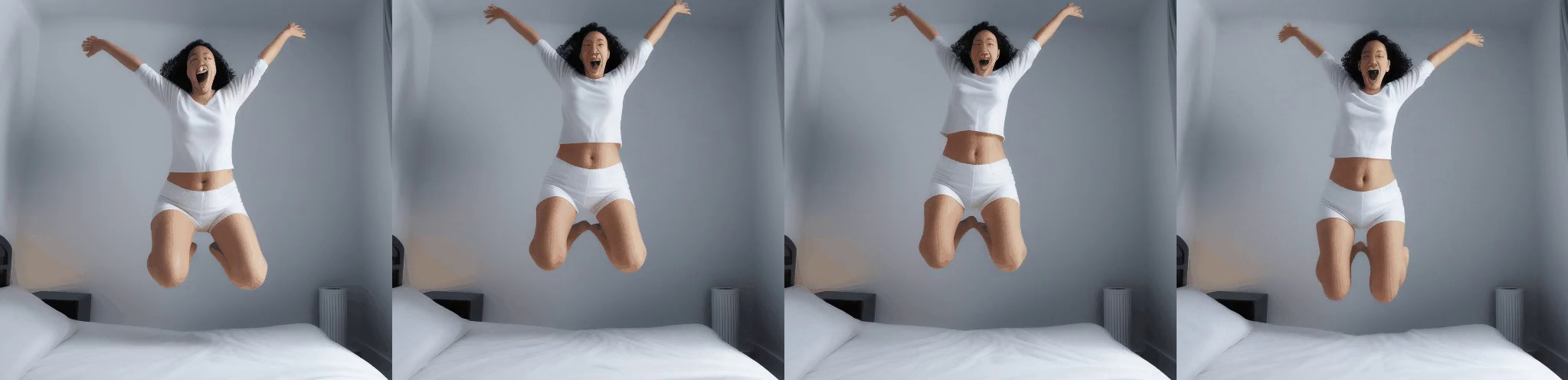}
    \caption{\textbf{Prompt:} A woman is jumping on a white bed. \textbf{\textcolor{red}{{Subject Dysmorphia}:}} The video depicts a woman jumping on a white bed. Over time, a hallucination effect manifests, leading to a dysmorphic transformation of the woman's face within the video.}
    \label{fig:td2}
\end{figure*}

\begin{figure*}
    \centering
    \includegraphics[width=\linewidth]{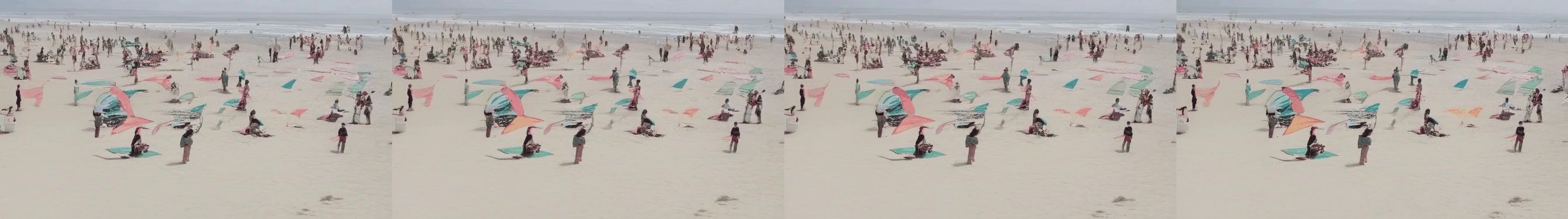}
    \caption{\textbf{Prompt:} A crowd of people standing on a beach flying kites. \textbf{\textcolor{red}{{Visual Incongruity}:}} Instead of being depicted in the sky as expected, the kites appear visually inconsistent, resembling objects embedded in the sand.}
    \label{fig:pi1}
\end{figure*}

\begin{figure*}
    \centering
    \includegraphics[width=\linewidth]{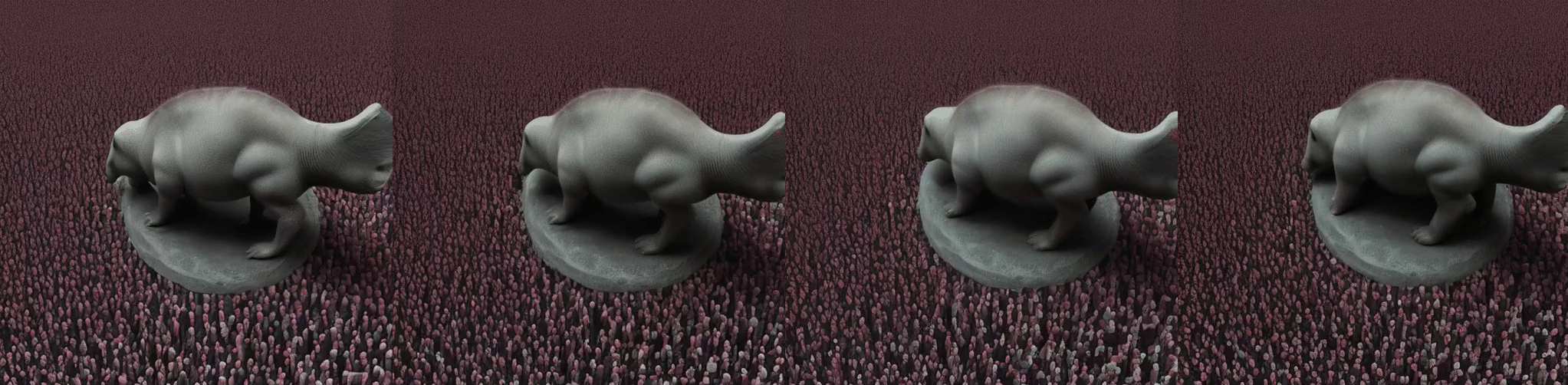}
    \caption{\textbf{Prompt:} a animal that is walking in a crowd of people \textbf{\textcolor{red}{{Visual Incongruity}:}} In the generated video, a stone statue of an animal is seen moving atop a vast crowd that appears to be composed of human heads. The statue's movement contrasts with its rigid, lifeless material, creating an unsettling effect. The generated video blurs the line between the inanimate and the living.}
    \label{fig:pi2}
\end{figure*}

\begin{enumerate}

\item \textbf{{Vanishing Subject (VS):}} See \cref{fig:vs1,fig:vs2}
\item \textbf{{Omission Error (OE):}} See \cref{fig:oe1,fig:oe2}
\item \textbf{{Numeric Variability (NV):}} See \cref{fig:nv1,fig:nv2}
\item \textbf{{Subject Dysmorphia (SD):}} See \cref{fig:td1,fig:td2}
\item \textbf{{Visual Incongruity (VI):}} See \cref{fig:pi1,fig:pi2}

\end{enumerate}

\end{document}